\documentclass[letterpaper]{article}
\usepackage{iccc} 
\usepackage{amsmath, amsthm}
\usepackage{enumitem}
\usepackage{times}
\usepackage{helvet}
\usepackage{courier}
\usepackage{graphicx} 
\usepackage{booktabs}
\usepackage{caption}
\usepackage{url}
\usepackage[table]{xcolor}
\usepackage[most]{tcolorbox}

\definecolor{blacktext}{rgb}{0, 0, 0}
\definecolor{lightorangebg}{HTML}{E8C8BA}
\definecolor{darkorange}{HTML}{E59D80}
\definecolor{lightbluebg}{HTML}{CADCEA}
\definecolor{darkblue}{HTML}{5DA6E9}
\colorlet{tblbg}   {lightbluebg} 
\colorlet{tblframe}{darkblue}
\colorlet{tblhead} {darkblue}

\title{\textsc{Spark}: A System for Scientifically Creative Idea Generation}
\author{Anonymous Authors}
\author{Aishik Sanyal \\
Spiral Works \\ 
aishik@spiralworks.ai
\And 
Samuel Schapiro \\
Univ.\ Illinois, Urbana-Champaign \\
Spiral Works \\
sam@spiralworks.ai \\
\And
Sumuk Shashidhar\\ 
Univ.\ Illinois, Urbana-Champaign \\ 
sumuks2@illinois.edu
\AND 
Royce Moon \\
University of Michigan \\ 
Spiral Works \\
royce@spiralworks.ai \\
\And
Lav R. Varshney \\ 
 Univ.\ Illinois, Urbana-Champaign\\
varshney@illinois.edu\\
\And Dilek Hakkani-Tur  \\
Univ.\ Illinois, Urbana-Champaign \\
dilek@illinois.edu
}
\begin{document} 
\maketitle
\begin{abstract}
\begin{quote}
Recently, large language models (LLMs) have shown promising abilities to generate novel research ideas in science, a direction which coincides with many foundational principles in computational creativity (CC). In light of these developments, we present an idea generation system named \textsc{Spark} that couples retrieval-augmented idea generation using LLMs with a reviewer model named \textsc{Judge} trained on 600K scientific reviews from OpenReview. Our work is both a system demonstration and intended to inspire other CC researchers to explore grounding the generation and evaluation of scientific ideas within foundational CC principles. To this end, we invite other researchers to explore the use of LLMs for idea generation and creative evaluations, and release the annotated dataset used to train \textsc{Judge}.\footnote{\url{https://huggingface.co/datasets/spiralworks/openreview_wildcard_2025_v2}}

\end{quote}
\end{abstract}

\section{Introduction}
Automating parts of the scientific process has long been a research goal in the field of computer science and artificial intelligence (AI) \cite{langley1977bacon,thagard1988computational,alphafold}, with computational creativity (CC) principles playing a crucial role in many parts of this endeavor. Crucially, the initial task of idea or hypothesis generation is an area that requires creative intuition but has historically been restricted to the ingenuity of human scientists.

Recent research, however, has demonstrated that large language models (LLMs) are capable of generating novel research ideas \cite{scimon,si2024can,yang2023large}. In fact, a handful of fully autonomous AI systems capable of writing full-length papers at near-human quality, or workshop-length papers that successfully pass through human peer review, have been developed
\cite{sakana2024,zochi2025}. In each of these systems, idea generation remains the crucial first step.

In this paper, we present a scientific idea generation system named \textsc{Spark} that generates novel scientific ideas and an evaluator model named \textsc{Judge}, trained on a 600K size dataset of expert human reviews from OpenReview, that can evaluate the creativity of \textsc{Spark}'s proposed ideas. Our goal is to ground recent developments in scientific idea generation with the foundational computational creativity principles that underpin much of their work, focusing on the first stage of idea generation. To facilitate reproducibility and community progress, we are releasing the full OpenReview dataset used to train our \textsc{Judge} model. Finally, to provide background context, we first start by summarizing recent developments in the use of LLMs for scientific discovery.

\section{Related Work}
Large language models (LLMs) have been increasingly explored as tools to support creative ideation in scientific research. Early efforts such as Meta’s Galactica utilized specialized scientific corpora to facilitate technical tasks like knowledge retrieval and hypothesis generation \cite{taylor2022galactica}. Despite strong benchmark performance, Galactica encountered significant issues with factual grounding and hallucinated references, highlighting the need for improved reliability.

Addressing grounding explicitly, SciPIP integrated semantic retrieval and citation awareness into idea generation, producing hypotheses informed by current literature but primarily evaluated within NLP domains \cite{scipip}. Extending this further, SciMON employed iterative refinement to systematically maximize the novelty of ideas, but their outputs often suffered from low practical feasibility and experimental utility \cite{scimon}.

Drawing inspiration from collaborative research processes, VirSci orchestrated multiple LLM agents within a virtual ecosystem to simulate team-based ideation, achieving measurable improvements in idea diversity and research-trend alignment \cite{wang2024virsci}. Nevertheless, like SciMON, VirSci primarily focused on novelty without directly optimizing for experimental usefulness.

Other efforts aimed at automating entire research cycles include Sakana AI’s AI Scientist, which autonomously generates and experimentally validates hypotheses but still requires human oversight for accuracy and consistency \cite{sakana2024}. Similarly, IntologyAI’s Zochi employed multi-agent collaboration to produce workshop-accepted research papers across domains, yet it involves significant human intervention and computational resources \cite{zochi2025}.

In this paper, we present \textsc{Spark}, a demonstration aimed at grounding scientific idea generation in foundational computational creativity principles. We illustrate this approach by integrating literature retrieval, idea generation, and simulated peer-review evaluation into a unified workflow. Additionally, we invite further exploration from the CC community by releasing the OpenReview dataset used to train the Judge model.

\section{Methods}

The \textsc{Spark} system integrates AI-driven components that support computational creativity in the generation of scientific ideas, utilizing specialized agent-based modules for the retrieval of literature and the synthesis, refinement and evaluation of ideas. Figure 1 illustrates the interaction between the \textsc{Spark} and \textsc{Xplor} systems.

\subsection{Literature Retrieval via \textsc{Xplor}}

Inspired by PaperQA \cite{lala2023paperqa}, \textsc{Xplor} handles literature retrieval and analysis tasks tailored specifically for the \textsc{Spark} system. It utilizes OpenAI embeddings, FAISS indexing, maximal marginal relevance (MMR), and a recursive, agent-based retrieval loop to systematically gather, analyze, and refine relevant research documents to later be used for idea generation.

\subsubsection{Text Embedding and Retrieval:}

\textsc{Xplor} employs OpenAI's \texttt{text-embedding-3-large} model \cite{openai_text_embedding_3_large_2024} to convert retrieved documents into 1536-dimensional semantic vectors. These embeddings allow accurate semantic comparison of research papers. FAISS \cite{douze2024faiss} efficiently indexes these vectors, enabling rapid retrieval through cosine similarity-based nearest-neighbor searches when provided with user-generated input queries.

\subsubsection{Relevance Scoring and Maximal Marginal Relevance:}

After an initial vector-based retrieval, each candidate text chunk is also evaluated using an LLM-based summary model. This model assigns relevance scores (1 to 10) to each chunk according to its relevance to the research query, complementing the earlier cosine similarity measures. Subsequently, \textsc{Xplor} applies maximal marginal relevance (MMR) to re-rank these scored chunks, a technique which prioritizes both relevance and diversity. This ensures that selected documents collectively provide comprehensive coverage with minimal redundancy.

\subsubsection{Recursive Retrieval and Evidence Gathering Loop:}

The retrieval process in \textsc{Xplor} is recursive and agent-driven. Initially, the system queries databases (e.g., arXiv, Google Scholar) using a programmable search engine, retrieving documents relevant to the initial question. First, documents are segmented into overlapping chunks and embedded. Then, an agentic loop iteratively refines the search by generating new, increasingly precise queries informed by intermediate summaries from an LLM response. This process continues until specific depth and coverage criteria, such as accumulating at least five evidence snippets from multiple sources, are met. 
\begin{figure}[ht]
  \centering
  \includegraphics[width=\columnwidth]{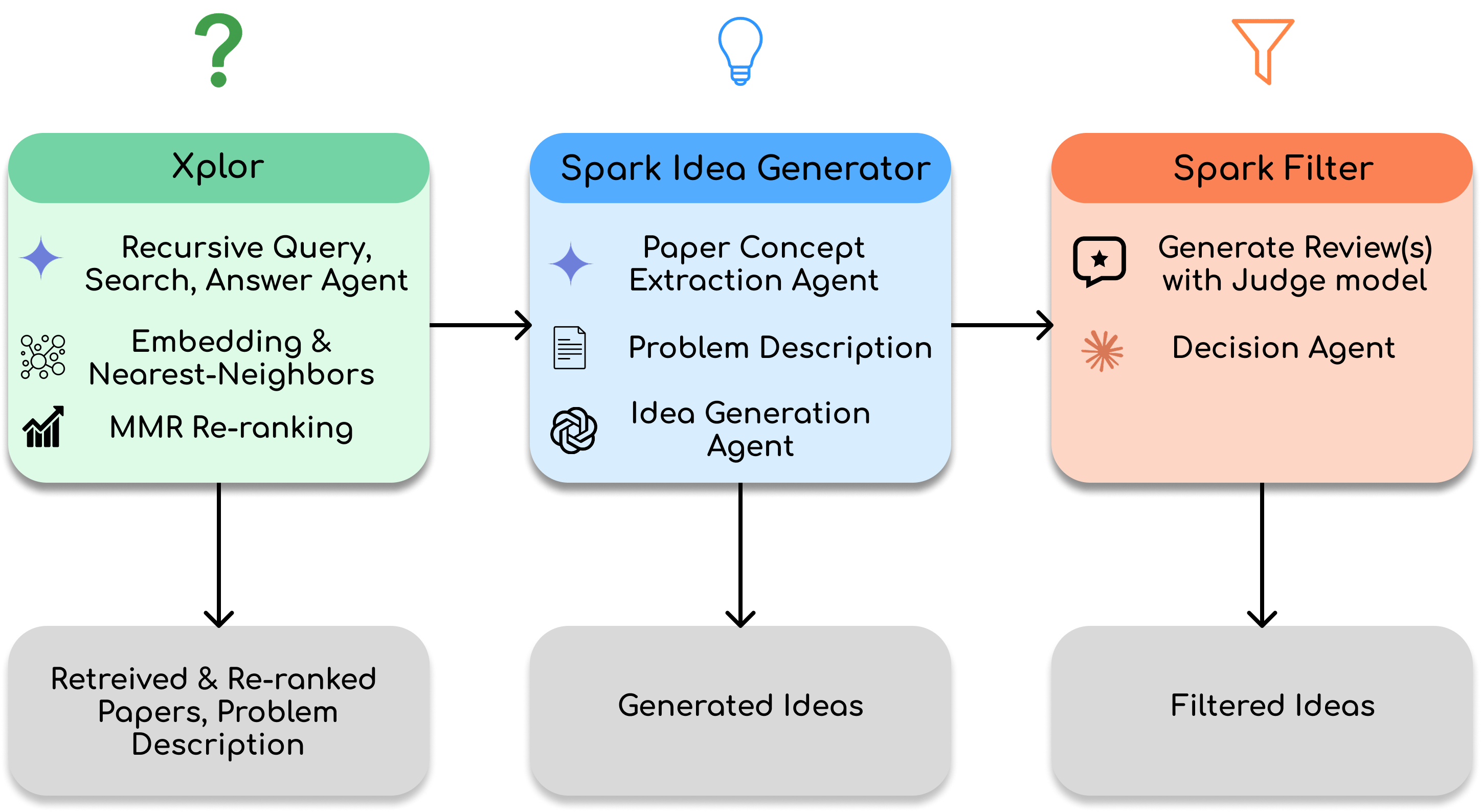} 
  \caption{\textsc{Spark}’s end‑to‑end pipeline: \textsc{Xplor} performs embedding‑based, recursive literature retrieval with MMR re‑ranking; the \textsc{Spark} Idea Generator extracts key concepts and uses chain‑of‑thought LLM prompting to synthesize seed research proposals; and the \textsc{Spark} Filter applies our supervised fine-tuned \textsc{judge} model and a decision agent for automated peer‑style critique and ranking.}
  \label{fig:idea}
\end{figure}
\textsc{Xplor} has two distinct operating modes: interactive and autonomous. In interactive mode, \textsc{Xplor} is directly prompted by the user query, whereas in autonomous mode, \textsc{Xplor} takes queries from the \textsc{Spark} Idea Generator agent to aid in the generation and refinement of ideas, as explained in the following section.

\subsection{Idea Generation with \textsc{Spark} Idea Generator}
The \textsc{Spark} Idea Generator constitutes the initial creative stage of the \textsc{Spark} system, formulating seed research proposals based on the earlier \textsc{Xplor} search results. The idea generation system identifies open research problems related to the query and then synthesizes concepts to generate structured, contextually relevant, and well-motivated ideas.
\begin{figure}[ht]
  \centering
  \scriptsize
  \begin{minipage}{\columnwidth}
    \textbf{A) Input:}
        \begin{tcolorbox}[colback=lightbluebg,coltext=blacktext,boxrule=0pt,arc=2mm,boxsep=1pt]
    \begin{itemize}\itemsep0pt
      \item \textbf{Problem:} “LLMs struggle with consistency”
      \item \textbf{Concepts:} [Chain-of-Thought (CoT), Knowledge Graphs (KGs), …]
      \item \textbf{Domain:} NLP‑LLMs
    \end{itemize}
    \end{tcolorbox}
    \vspace{0.3\baselineskip}

    \textbf{B) Idea Generation Prompt:}
    \begin{tcolorbox}[colback=lightbluebg,coltext=blacktext,boxrule=0pt,arc=2mm,boxsep=1pt]
\begin{verbatim}
Generate a novel research idea with step-by-step 
reasoning based on the provided input. Output must
follow the following template.
\end{verbatim}
    \end{tcolorbox}
    \vspace{0.3\baselineskip}

    \textbf{C) LLM Response:}
    \begin{tcolorbox}[colback=lightbluebg,coltext=blacktext,boxrule=0pt,arc=2mm,boxsep=1pt]
\begin{verbatim}
{ "input_concepts":["CoT","KGs"],
  "new_concepts":["Entity linking"],
  "plan":"Analyze. Wait. Try a new perspective. 
          Propose KG lookups",
  "title":"KG-CoT for Fact Consistency",
  "abstract":"Integrate KGs for dynamic grounding" }
\end{verbatim}
    \end{tcolorbox}
    \vspace{0.3\baselineskip}

    \textbf{D) Generated Idea:} 
    Title and Abstract from the structured response in (C).
  \end{minipage}
  \caption{Spark Idea Generation pipeline: (A) Inputs, (B) Idea Generation Prompt, (C) LLM Response, (D) Generated Idea.}
  \label{fig:spark_idea_gen_prompt}
\end{figure}
\begin{figure}[ht]
  \centering
  \scriptsize
  \begin{minipage}{\columnwidth}
    
    \textbf{A) Input:}
\begin{tcolorbox}[colback=lightorangebg,boxrule=0pt,arc=2mm,boxsep=1pt]    
    \begin{itemize}\itemsep0pt
      \item \textbf{Title:} “KG-CoT for Fact Consistency…”
      \item \textbf{Abstract:} “Integrate KGs for dynamic grounding…”
      \item \textbf{Reviews:}
      \begin{itemize}\itemsep0pt
        \item Review 1: “Critique: Promising KG-CoT...”
        \item Review 2: “Strengths: Addresses key issue...”
        \item Review 3: “Concerns: Novelty vs prior…”
      \end{itemize}
    \end{itemize}
  \end{tcolorbox}

\textbf{B) Filter Prompt:}
    \begin{tcolorbox}[colback=lightorangebg,boxrule=0pt,arc=2mm,boxsep=1pt]  
    \begin{itemize}\itemsep0pt
      \item Generate your decision based on the provided input Idea and Reviews.
      \item Output must follow the following template:
      \begin{itemize}\itemsep0pt
        \item \texttt{"Decision reasoning"} (string)
        \item \texttt{"Decision"} (ACCEPT/REJECT)
        \item \texttt{"Utility"} (float 0–1)
      \end{itemize}
    \end{itemize}
\end{tcolorbox}
  
    \textbf{C) LLM Response:}
\begin{tcolorbox}[colback=lightorangebg,boxrule=0pt,arc=2mm,boxsep=1pt]
\begin{verbatim}
{ "Decision reasoning": "Detailed critique…",
  "Decision": "REJECT",
  "Utility": 0.35 }
\end{verbatim}
    \end{tcolorbox}
\textbf{D) Generated Decision:} 
    \texttt{REJECT}.
  
  \end{minipage}
  
\caption{Spark Filter pipeline: (A) Input, (B) Filter Prompt, (C) LLM Response, (D) Generated Decision.}
  \label{fig:spark_filter_prompt}
\end{figure}

\begin{figure*}[ht]
  \centering
  \footnotesize   

  \begin{tcolorbox}[
    enhanced,
    colback=tblbg,
    colframe=tblframe,
    colbacktitle=tblhead,
    coltitle=white,
    fonttitle=\bfseries,
    boxrule=0.8pt, arc=2mm,
    left=4pt, right=4pt, top=4pt, bottom=4pt,
    width=\linewidth,
    title={Entropy Reduction and Intrinsic Motivation in AI Music Creation},
  ]
    This research explores the simulation of human music creation using AI, focusing on incorporating intrinsic motivation and addressing the potential impact on human artists. We propose a system based on a Transformer network trained with reinforcement learning to generate music. The core idea is to model musical creativity as a process of 'restructuring high psychological entropy material,' where entropy represents the complexity and unpredictability of musical elements. The AI is trained with an intrinsic reward function that encourages it to reduce the entropy of its generated music, quantified using Kullback-Leibler Divergence between probability distributions of musical features before and after transformations. This reward is shaped to encourage exploration and discovery of new musical structures. The system's performance will be evaluated by comparing its generated music to human-composed music using perceptual audio features (MFCCs, chromagrams) and subjective ratings through metrics like cosine similarity and t-SNE visualization. Furthermore, we will conduct a survey to assess how exposure to the AI-generated music affects human artists' motivation and creative output, measuring changes in their creative output and self-reported motivation levels. This research aims to provide insights into the computational modeling of creativity and address the ethical and societal implications of AI in the arts. It seeks to understand how AI can be a tool that both generates and inspires music.
  \end{tcolorbox}

  \caption{A research idea proposed by \textsc{Spark}, using context provided by \textsc{Xplor} about the use of LLMs for scientific creativity.}
  \label{tab:two_creativity_ideas}
\end{figure*}

\textbf{Concept Extraction and Problem Synthesis:} Initially, a user provided research question guides \textsc{Xplor} to recursively retrieve relevant papers and identify open research problems. The concepts and open problems from these papers are then used to construct a problem-driven research prompt (Figure 2, A-B).

\textbf{Structured Idea Formulation:} Using the constructed prompt, an LLM agent generates a preliminary research proposal comprising a structured reasoning plan, novel concepts, and a concise title and abstract, guided to incorporate relevant domain knowledge and maintain logical coherence (Figure 2, C-D). This output (as illustrated in Figure 4) forms the basis for high-quality research ideas already contextualized within existing literature and ready to be explored by users of the system. If further refinement is needed, the system adaptively uses \textsc{Xplor} to retrieve more relevant literature to address a specific area of improvement.

\subsection{Idea Evaluation with \textsc{Spark} Filter}
\textsc{Spark} Filter acts as the evaluative module within the \textsc{Spark} framework, simulating a scalable peer review process to assess proposed research ideas' viability and quality.

\textbf{Review Generation:} In this phase, each idea—represented by its title and abstract—is processed by our supervised fine-tuned (SFT) \textsc{Judge} model trained on OpenReview peer review corpora. This model generates multiple critiques per idea, highlighting strengths, limitations, and areas for improvement, simulating the diverse perspectives that might be offered during the scientific peer review process.

\textbf{Decision Synthesis:} Subsequently, a separate LLM integrates these reviews to produce a final binary classification—\texttt{ACCEPT} or \texttt{REJECT}—of the idea, an explanation of the decision, and utility score for the idea (Figure 3, A-D).

This pipeline has successfully produced and filtered over 10,000 AI-focused research ideas, demonstrating its scalability for large-scale computational creativity tasks. In future work, it is worth exploring techniques to enhance the creativity and diversity of \textsc{Spark}'s generated ideas \cite{si2024can,scimon}.

\subsection{Evaluator Training: The \textsc{Judge} Model}

Assessing the creativity and novelty of AI‑generated scientific ideas demands a rigorous evaluator. Off‑the‑shelf LLMs, trained via supervised fine‑tuning (SFT) or reinforcement learning from human feedback (RLHF), tend to prioritize agreeableness over critical scrutiny \cite{ouyang2022training}, making them ill‑suited to judge nascent research proposals. To overcome this, we introduce \textsc{Judge}, a purpose‑built model trained to critique scientific ideas according to established academic standards.

\subsubsection{Data Acquisition and Preprocessing}
We sourced training data from OpenReview’s publicly available submissions, peer reviews, and meta‑reviews. An automated pipeline harvested abstract–review pairs to form the core dataset. Although we initially explored mapping full papers to their reviews, this blended conceptual appraisal with evaluations of implementation and results. Focusing on abstracts alone isolates and sharpens the model’s assessment of underlying ideas.  

A further challenge arises from the content of abstracts themselves. Authors often highlight strong empirical results, which can positively bias reviewer scores irrespective of the underlying idea's intrinsic novelty or merit. To mitigate this inherent bias, we introduced a data transformation step. We define an ``idea abstract" ($\mathcal{A}_{idea}$) as a version of the original abstract ($\mathcal{A}_{orig}$) that deliberately omits specific results, implementation details, and performance metrics, while preserving the core problem statement, proposed approach, and claimed novelty. Similarly, an ``idea review" ($\mathcal{R}_{idea}$) represents a critique focused exclusively on these conceptual aspects, abstracted away from the empirical outcomes presented in the original review ($\mathcal{R}_{orig}$).

To generate this refined dataset at scale, we employed DeepSeek-V3 as an automated annotator \cite{deepseek}. Given an $(\mathcal{A}_{orig}, \mathcal{R}_{orig})$ pair, DeepSeek-V3 was prompted to perform two separate transformations: generating $\mathcal{A}_{idea}$ from $\mathcal{A}_{orig}$, and generating $\mathcal{R}_{idea}$ from $\mathcal{R}_{orig}$. The choice of DeepSeek-V3, an open-weight model, over potentially stronger proprietary models was made explicitly to enhance the reproducibility of our data generation process.

\subsubsection{Multi-Task Training Framework}
The \textsc{Judge} model was trained using a multi-task objective designed to capture both evaluative and generative aspects of scientific discourse. We formulated four distinct tasks, using specialized input formats and system prompts to differentiate them.
\begin{enumerate}
    \item \textbf{Original Review Prediction:} Mapping an original abstract to its corresponding original review, $f_{eval}: \mathcal{A}_{orig} \rightarrow \mathcal{R}_{orig}$.
    \item \textbf{Idea Review Prediction:} Mapping an idea abstract to its corresponding idea review, $f_{idea\_eval}: \mathcal{A}_{idea} \rightarrow \mathcal{R}_{idea}$.
    \item \textbf{Original Abstract Generation:} Mapping an original review back to its corresponding original abstract, $f_{gen}: \mathcal{R}_{orig} \rightarrow \mathcal{A}_{orig}$.
    \item \textbf{Idea Abstract Generation:} Mapping an idea review back to its corresponding idea abstract, $f_{idea\_gen}: \mathcal{R}_{idea} \rightarrow \mathcal{A}_{idea}$.
\end{enumerate}
Training encompassed these forward (evaluation) and backward (generation) mappings concurrently. This multi-task approach encourages the model to develop a deeper, bidirectional understanding of the relationship between a research idea and its critical appraisal.

The model was trained via continued pretraining on this structured dataset, employing low-rank adaptation (LoRA) to efficiently update the base model's weights \cite{lora}. The training objective was to maximize the log-likelihood of the target sequence (either a review or an abstract, depending on the task) given the input sequence, formally represented as maximizing $\mathcal{L}(\theta) = \sum_{(x, y)} \log P(y | x; \theta)$, where $(x, y)$ represents an input-output pair from one of the four tasks, and $\theta$ denotes the model parameters.

\subsubsection{Evaluation and Model Scaling}
To assess the performance of \textsc{Judge}, we used a held-out temporal partition of the OpenReview dataset, comprising reviews and associated metadata submitted after a predefined cutoff date (e.g., post-October 2024 or from 2025 onwards). Evaluation metrics included the quality of generated reviews (assessed qualitatively and via RMSE of the final review scores in aggregate) and the accuracy of predicting explicit review scores where available.

To investigate the impact of model scale on evaluation capabilities, we trained two versions of \textsc{Judge}:

\begin{itemize}
    \item \texttt{wintermute-tiny}: Based on the Qwen-7B model.
    \item \texttt{wintermute-medium}: Based on the Qwen-72B model.
\end{itemize}

This provides a roughly 10$\times$ increase in parameter count, allowing for analysis of scaling effects. Empirical validation confirmed that both \texttt{wintermute-tiny} and \texttt{wintermute-medium} consistently outperform their respective base models and other contemporary state-of-the-art LLMs on the designated evaluation tasks, demonstrating the effectiveness of our specialized training approach.
\section{Discussion and Limitations}
The \textsc{Spark} prototype integrates literature retrieval, multi-agent idea generation, and a simulated peer-review filter into a single workflow intended to illustrate how computational-creativity principles can be operationalized in scientific ideation.  Each stage is deliberately modular: \textsc{Xplor} provides a retrieval layer, the Idea Generator converts retrieved context into candidate hypotheses, and the Filter simulates a scientific peer review step.  By decoupling these stages, the system offers a test bed for exploring how generation and evaluation strategies interact within the context of scientific idea generation.

Nevertheless, several design choices merit further exploration and refinement.  First, the current filter operates only once per idea; future studies could examine closed-loop variants that feed reviewer feedback back into the generator.  Second, while \textsc{Xplor}’s retrieval is similarity-based, structured knowledge sources (e.g.\ ontologies) might support deeper analogical reasoning.  Third, the present peer-review simulation produces textual critiques but no explicit explanatory traces; richer explanation outputs could help users understand the basis of automatic judgments.  Finally, \textsc{Spark} has been evaluated only in a controlled setting confined to idea generation within the field of AI, but to understand its abilities across other disciplines,  studies in additional scientific fields are required.

In summary, we hope this proof-of-concept can serve as a guide for future work in this area, encouraging comparative studies of retrieval methods, generation prompts, reviewer-agent designs, and creativity evaluation strategies using our system as a foundation. To facilitate community experimentation, we have released the annotated dataset, consisting of papers and reviews, used to train the \textsc{Judge} model. 

\section{Looking Ahead: Computational Creativity and Scientific Idea Generation}
The use of AI systems to generate scientific ideas raises important considerations that the CC community is well-poised to address. We state some open questions on this topic below, with the hope of inspiring other CC researchers to begin exploring these issues in future work.

\subsection{Creative Criteria for Ideas Versus Papers} Creativity is typically measured by three dimensions: novelty, utility, and surprise \cite{boden2004creative,creativity_in_science,maher2010evaluating}. While the utility of a scientific paper is mostly judged by its future impact—whether that be citation count, accolades like best paper awards, and other retrospective metrics—idea generation and evaluation systems such as ours are tasked with finding high-quality directions that merit further exploration before their research outcomes are ever observed. How should CC researchers approach utility evaluations for scientific ideas? We believe a combination of theory-guided and data-driven approaches may be most fruitful here, with our dataset release helping facilitate the latter.

\subsection{Transformative Idea Generation}
Similarly, creative processes are typically divided into three types: combinatorial, exploratory, and transformational \cite{boden2004creative}. Our idea generation system relies on concept extraction and problem synthesis to construct well-motivated and contextually relevant ideas, often through the combination of existing concepts, as outlined in Figure~\ref{fig:spark_idea_gen_prompt}. Nevertheless, many historically groundbreaking discoveries have been transformational, actively restructuring elements of the conceptual spaces afforded by their scientific field. Can AI systems be used to discover transformational scientific ideas? Considering the historical impact transformational creativity has had on science and society at large, we believe this direction merits considerable future exploration.

\section{Conclusion}
This paper presented \textsc{Spark} as a system demonstration that unifies a literature retrieval engine, an LLM-driven idea-generation module, and a peer-review-inspired filter. The implementation and discussion in this work are offered as a concrete step towards grounding scientific idea generation in foundational computational creativity principles. By illustrating how creative generation and evaluation can be combined, we aim to inspire new directions in the computational creativity community around the generation, evaluation, and feasibility of scientific idea generation using LLM systems. We anticipate that the framework, data resources, and observations reported here will serve as a starting point for future work exploring the relationship between computational creativity and scientific idea generation.

\appendix

\bibliographystyle{iccc}
\bibliography{iccc}

\end{document}